# Online Tracking of Skin Colour Regions Against a Complex Background

S. Basu*, S. Chakraborty[+], K. Mukherjee*, S. K. Pandit*
* Computer Sc. & Engg. Dept., [+] Electronics & Comm. Engg. Dept.,
MCKV Institute of Engineering, 243,G T Road (North); Liluah, Howrah-711204, India.
Email: {subhadip8@yahoo.com}

*Abstract: Online tracking of human activity against a complex background is a challenging task for many applications. In this paper, we have developed a robust technique for localizing skin colour regions from unconstrained image frames. A simple and fast segmentation algorithm is used to train a multiplayer perceptron (MLP) for detection of skin colours. Stepper motors are synchronized with the MLP to track the movement of the skin colour regions.*

## 1. Introduction

Online recognition of skin colour regions is an important task for many applications. Applications such as face recognition, interpretation of gestures, video databases, virtual reality interfaces, security monitoring all have in common the need to track and interpret human activities. The ability to localize and track people's face and hands against a complex background is therefore an important visual problem for the researchers.

To address this need, we have developed a real-time system for tracking of human activities based on skin colour. The single rotating camera tracks the facial and hand movements based on the mean position (x, y) of the skin colour regions in the image plane.

The problem of face localization and recognition has long been an active area of interest for the researchers in computer vision [1], [2], [3], [4], [5]. In [1], R. Féraud *et al.* proposed a fast and simple face localization technique based on motion and skin colour filters. C.R. Wren *et.al.* [6] have used the colour and spatial similarity properties to form connected regions in which all pixels have similar image properties. However, both the techniques use high-end, costly gadgets and complex computation techniques for segmentation and tracking of human activities. Y. Yacoob *et.al.* [7] have used a technique of computation of the translation of a location of interest (facial muscles) in successive image frames to identify the facial expressions. But, the work does not concentrate on the possibility of a complex background. Rolf P. Würtz , in his work on object recognition [8], have used Gabor wavelet features for sampling the image, amplitude thresholding, background suppression and image representation. The computational complexities of these algorithms are high for a real-time application. For recognition of hand gestures, feature based approaches are used in [9], [10], [11]. In our previous work [12], we had considered hand gestures against an uniform contrasting background. However, for online localization and tracking of body movements in an unconstrained environment, the object of interest must be isolated from a complex background quickly and effectively.

In our work, we have used a simple segmentation technique based on the RGB intensity values of an image frame. A multi-layer perceptron (MLP) is then trained with the mean RGB values of the skin colour segments and the trained network is used to detect skin colours from the real-time image frames. The complete work is divided into the following modules, *viz.* interfacing with the camera, image segmentation, design of the MLP and image tracking.

## 2. Interfacing with the Camera

In our work, we have used a Logitech® QuickCam® Express camera and its software development kit [13] for capturing real-time video information. 24 bit colour image sequences at 5 frames per second, of size 320x240 pixels are considered for experimentation. The video notification callback enables video frames to be transferred from the video control to the calling application. The notification message is received during video streaming. Video streaming allows direct access to camera data through the notification message. This allows applications to access video data directly in the fastest possible manner.

### 3. Image Segmentation

The term segmentation means to divide or segment an image into a number of regions, each of which is reasonably uniform in some characteristics. A region can be defined as a collection of adjacent pixels that are similar in some way, such as brightness, color or local visual texture.

Table 1. Region-labeling algorithm

| |
|---|
| Initialize label, L ← -1 |
| For all (x, y) scan the image from left to right and top to bottom |
| IF RGB(x, y)>=0 then |
| Insert pixel (x, y) to List |
| While List is not empty |
| Remove a pixel (s, t) from List |
| For each 4-neighbour (u, v) of (s, t) |
| If (u, v) is unlabelled and |RGB(u, v) - RGB(x, y)| < η then |
| Insert (u, v) to List |
| End For |
| End While |
| L ← L – 1 |
| End For |

In our work, we have used the RGB values of each pixel for this segmentation process. A simple 4-connected region-labeling algorithm [14] (as shown in Table. 1) is used to segment the colour image frames of size 320x240.

The function RGB(x, y) in Table 1, represents the red, green and blue intensity values of the pixel respectively and η is a threshold on changes in intensity values for identifying neighbouring pixels within a region. The time complexity of this algorithm is of O(n), where n is the number of pixels in the image frame. The algorithm is fast and easy to implement in real-time. The threshold η can be adjusted to tune the sensitivity of the segmentation process. Lower the value of η, more sensitive is the algorithm towards changes in RGB intensity values.

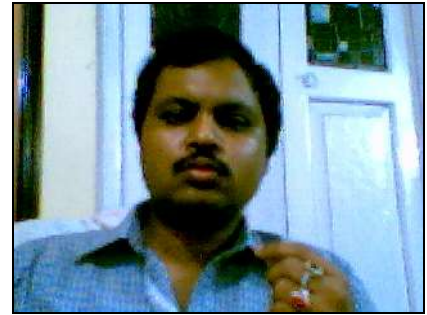
(a)

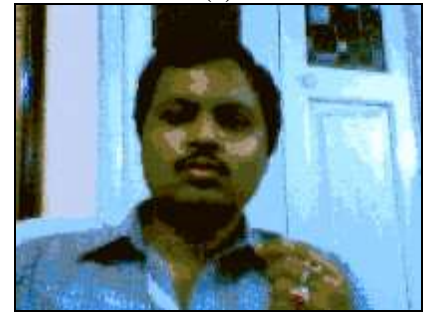
(b)

Fig. 1. A sample 24 bit colour image frame of size 320x240 (a) and the result after segmentation with η = 28 (b)

### 4. Design of the MLP

We have used a multilayer perceptron with back propagation learning algorithm for detection of the skin colour from the image frames. Red, green and blue intensity values of each pixel are used as basic features. From sample segmented images of different persons, we have manually selected skin colour regions and stored in a skin colour image database (as shown in fig. 2). Unique RGB values, as shown in table 2, are identified from these skin database images. A 3-3-1 network is trained with 50 training samples of unique RGB skin colour samples. The network is trained for 200 iterations with learning rate as 0.6 and acceleration factor as 0.7.

During the online image analysis, each image frame is first segmented using the algorithm discussed earlier. The average RGB

values of each unique label are fed to the neural network for testing of skin colour. If the output of the network is more than a pre-defined threshold, ρ, then the label representing the region is identified as skin colour region.

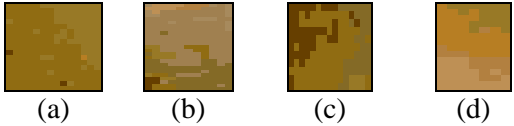

(a)　　　(b)　　　(c)　　　(d)
Fig. 2. Sample skin colour database images

Table 2. Sample RGB values generated from the skin database

| Red | Green | Blue |
|---|---|---|
| 35 | 126 | 183 |
| 85 | 144 | 190 |
| 64 | 128 | 178 |
| 80 | 128 | 160 |
| 38 | 106 | 132 |
| 38 | 121 | 152 |
| 18 | 108 | 144 |
| 0 | 63 | 102 |

## 5. Image Tracking

For tracking and localization of skin colours in the image frames, we have used stepper motors and the interfacing circuits (as shown in fig. 3). The interfacing circuit is designed using the ULN-2003 chip (stepper motor controller) and 74LS374 chip (Latch). 5 pin stepper motors are used to rotate the camera in horizontal and vertical directions.

The camera is mounted on the horizontal stepper motor. While analyzing the online image frames, the MLP identifies the skin colour regions from the image frame. The mean ($\mu_x$, $\mu_y$) of these skin pixels are then computed. The displacement of mean ($\mu_x$, $\mu_y$) is computed from center of the image (160, 120). The displacement values, ($|\mu_x -160|$, $|\mu_y -120|$), are fed to the stepper motor controlling module. Stepper motors, in both horizontal and vertical directions, rotate with unit angular movements to minimize the displacement values ($|\mu_x -160|$, $|\mu_y -120|$). In ideal case ($\mu_x$, $\mu_y$) coincides with (160,120) in real-time.

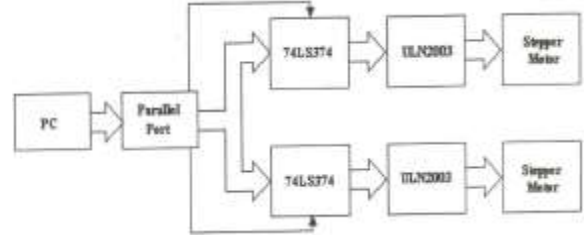

Fig. 3. Block diagram of the stepper motor controlling circuit.

## 6. Results and Discussion

In this work, we have designed a simple, low cost solution for online tracking of human activities based on skin colours. The output of a sample image frame is shown in fig. 5 (a-b). The technique gives reasonably good performance with a low-resolution camera. We have used a PIII 500MHz machine with 192MB primary memory for analysis of video frames. With a refresh rate of 5fps, the system shows some flickering effects. With machines of higher configuration, the system performance improves significantly. For the tracking system, we have tested the performance separately using stepper motors for horizontal and vertical directions. Higher configuration machines, with faster processing capabilities, are required for integration of both the stepper motors to the system.

In our future work, we plan to incorporate a more efficient movement detector with motion tracking systems and domain specific information. High-resolution camera with faster processing machines can improve the system performance significantly.

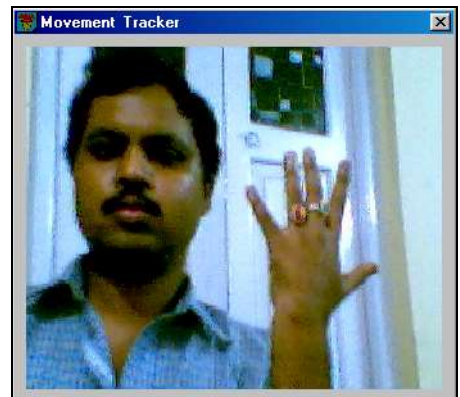

(a)

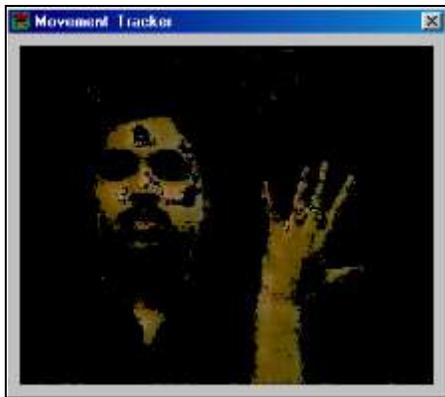
(b)

Fig. 4. A sample image frame (a) and the output after detection of skin colour regions(b).

**Acknowledgement**
Authors are thankful to the authorities of MCKV Institute of Engineering for kindly permitting to carry on the research work.